\newif\ifconf\conffalse 

\ifconf
\documentclass[oribibl,envcountsame]{llncs}
\usepackage{latexsym}
\else

\documentclass[12pt,twoside]{article}
\usepackage{latexsym}

\fi

\def\,{\mskip 3mu} \def\>{\mskip 4mu plus 2mu minus 4mu} \def\;{\mskip 5mu plus 5mu} \def\!{\mskip-3mu}
\def\dispmuskip{\thinmuskip= 3mu plus 0mu minus 2mu \medmuskip=  4mu plus 2mu minus 2mu \thickmuskip=5mu plus 5mu minus 2mu}
\def\textmuskip{\thinmuskip= 0mu                    \medmuskip=  1mu plus 1mu minus 1mu \thickmuskip=2mu plus 3mu minus 1mu}
\textmuskip
\def\eqsp{\vspace{0ex}}
\def\beq{\dispmuskip\eqsp\begin{equation}}    \def\eeq{\eqsp\end{equation}\textmuskip}
\def\beqn{\dispmuskip\eqsp\begin{displaymath}}\def\eeqn{\eqsp\end{displaymath}\textmuskip}
\def\bqa{\dispmuskip\eqsp\begin{eqnarray}}    \def\eqa{\eqsp\end{eqnarray}\textmuskip}
\def\bqan{\dispmuskip\eqsp\begin{eqnarray*}}  \def\eqan{\eqsp\end{eqnarray*}\textmuskip}

\ifconf\else
\newtheorem{theorem}{Theorem}
\newtheorem{corollary}[theorem]{Corollary}
\newtheorem{lemma}[theorem]{Lemma}
\newtheorem{definition}[theorem]{Definition}
\newtheorem{proposition}[theorem]{Proposition}

\newenvironment{keywords}{\centerline{\bf\small
Keywords}\vspace{-1ex}\begin{quote}\small}{\par\end{quote}\vskip
1ex}
\fi

\def\ftheorem#1#2#3{\begin{theorem}[#2]\label{#1} #3 \end{theorem} }

\def\flemma#1#2#3{\begin{lemma}[#2]\label{#1} #3 \end{lemma} }
\def\fdefinition#1#2#3{\begin{definition}[#2]\label{#1} #3 \end{definition} }
\def\fproposition#1#2#3{\begin{proposition}[#2]\label{#1} #3 \end{proposition} }

\def\paragraph#1{\vspace{1ex}\noindent{\bf{#1.}}}

\def\toinfty#1{\stackrel{#1\to\infty}{\longrightarrow}}
\def\nq{\hspace{-1em}}
\def\qed{\hspace*{\fill}$\Box\quad$\vspace{1ex plus 0.5ex minus 0.5ex}}
\def\odt{{\textstyle{1\over 2}}}
\def\eps{\varepsilon}                   
\def\epstr{\epsilon}                    
\def\qmbox#1{{\quad\mbox{#1}\quad}}
\def\leqa{\tabcolsep=0.5ex \begin{tabular}{c} \\[-4ex] $\scriptstyle\hspace{-0.2ex}+$ \\[-1.3ex] $\leq$ \end{tabular}}

\def\eqm{\stackrel\times=}             
\def\leqm{\tabcolsep=0.5ex \begin{tabular}{c} \\[-4ex] $\scriptstyle\hspace{-0.2ex}\times$ \\[-1.3ex] $\leq$ \end{tabular}}
\def\geqm{\tabcolsep=0.5ex \begin{tabular}{c} \\[-4ex] $\scriptstyle\hspace{0.1ex}\times$  \\[-1.3ex] $\geq$ \end{tabular}}
\def\v{}
\def\l{{\ell}}                          
\def\M{{\cal M}}                        
\def\X{{\cal X}}                        
\def\S{{\cal S}}                        

\def\E{{\bf E}}                         
\def\P{{\bf P}}                         
\def\B{\{0,1\}}                        

\def\MM{M}                              
\def\e{{\rm e}}                        
\def\a{\alpha}
\def\o{\omega}

\ifconf\def\Set#1{{\if#1Q{I\:\!\!\!\!#1}\else\if#1Z{Z\!\!\!Z}\else{I\!\!#1}\fi\fi}}
\else\def\Set#1{{\if#1Q{I\;\!\!\!\!\!#1}\else\if#1Z{Z\!\!\!Z}\else{I\!\!#1}\fi\fi}}
\fi
\def\lb{\log}

\def\text#1{\mbox{\scriptsize{#1}}}

\begin{document}

\ifconf

\title{Universal Convergence of Semimeasures \\ on Individual Random Sequences}

\author{Marcus Hutter\inst{1} \and Andrej Muchnik\inst{2}}
\authorrunning{Marcus Hutter and Andrej Muchnik}
\institute{IDSIA, Galleria 2, CH-6928 Manno-Lugano, Switzerland%
\footnote[0]{This work was partially supported by the
Swiss National Science Foundation (SNF grant 2100-67712.02) and
the Russian Foundation for Basic Research
(RFBR grants N04-01-00427 and N02-01-22001).}
\\
\email{\hspace{1.3ex}marcus@idsia.ch, \hspace{4.2ex} http://www.idsia.ch/$^{_{_\sim}}\!$marcus}
\and
Institute of New Technologies, 10 Nizhnyaya Radischewskaya \\
\hspace{2ex}Moscow 109004, Russia, \hspace{7.5ex} \email{muchnik@lpcs.math.msu.ru}
}
\maketitle

\else

\title{\vskip -25mm\normalsize\sc Technical Report \hfill IDSIA-14-04
\vskip 2mm\bf\LARGE\hrule height5pt \vskip 3mm
\sc\centerline{Universal Convergence of Semimeasures}
    on Individual Random Sequences%
\thanks{This work was partially supported by the Swiss National Science Foundation
(SNF grant 2100-67712.02) and the Russian Foundation for Basic Research (RFBR grants N04-01-00427 and N02-01-22001).}
\vskip 4mm \hrule height2pt \vskip 0mm}
\author{
{\bf Marcus Hutter}\\[3mm]
\normalsize IDSIA, Galleria 2, CH-6928\ Manno-Lugano, Switzerland\\
\normalsize marcus@idsia.ch \hspace{8.5ex} http://www.idsia.ch/$^{_{_\sim}}\!$marcus\\[5mm]
{\bf Andrej Muchnik}\\[3mm]
\normalsize Institute of New Technologies, 10 Nizhnyaya Radischewskaya\\
\normalsize Moscow 109004, Russia \hfill
\normalsize muchnik@lpcs.math.msu.ru 
}
\maketitle

\fi

\ifconf\else\vspace{-5mm}\fi
\begin{abstract}
Solomonoff's central result on induction is that the posterior of
a universal semimeasure $\MM$ converges rapidly and with
probability 1 to the true sequence generating posterior $\mu$, if
the latter is computable. Hence, $M$ is eligible as a universal
sequence predictor in case of unknown $\mu$. Despite some nearby
results and proofs in the literature, the stronger result of
convergence for all (Martin-L{\"o}f) random sequences remained
open. Such a convergence result would be particularly interesting
and natural, since randomness can be defined in terms of $\MM$
itself. We show that there are universal semimeasures $M$ which do
not converge for all random sequences, i.e.\ we give a partial
negative answer to the open problem. We also provide a positive
answer for some non-universal semimeasures. We define the
incomputable measure $D$ as a mixture over all computable measures
and the enumerable semimeasure $W$ as a mixture over all
enumerable nearly-measures. We show that $W$ converges to $D$ and
$D$ to $\mu$ on all random sequences. The Hellinger distance
measuring closeness of two distributions plays a central
role.
\end{abstract}

\ifconf\else
\fi

\ifconf\else\pagebreak\fi
\section{Introduction}\label{secIntro}

A sequence prediction task is defined as to predict the next
symbol $x_n$ from an observed sequence $x=x_1...x_{n-1}$. The key
concept to attack general prediction problems is Occam's razor,
and to a less extent Epicurus' principle of multiple explanations.
The former/latter may be interpreted as to keep the simplest/all
theories consistent with the observations $x_1...x_{n-1}$ and to
use these theories to predict $x_n$. Solomonoff
\cite{Solomonoff:64,Solomonoff:78} formalized and combined both
principles in his universal prior $M$ which assigns high/low
probability to simple/complex environments $x$, hence implementing
Occam and Epicurus. Formally it is a mixture of all enumerable
semimeasures. An abstract characterization of $M$ by Levin
\cite{Zvonkin:70} is that $M$ is a universal enumerable
semimeasure in the sense that it multiplicatively dominates all
enumerable semimeasures.

Solomonoff's \cite{Solomonoff:78} central result is that if the
probability $\mu(x_n|x_1...x_{n-1})$ of observing $x_n$ at time
$n$, given past observations $x_1...x_{n-1}$ is a computable
function, then the universal posterior $M_n:=M(x_n|x_1...x_{n-1})$
converges (rapidly!) {\em with $\mu$-probability 1} (w.p.1) for
$n\to\infty$ to the true posterior
$\mu_n:=\mu(x_n|x_1...x_{n-1})$, hence $M$ represents a universal
predictor in case of unknown ``true'' distribution $\mu$.
Convergence of $M_n$ to $\mu_n$ w.p.1 tells us that $M_n$ is close
to $\mu_n$ for sufficiently large $n$ for almost all sequences
$x_1x_2...$. It says nothing about whether convergence is true for
any {\em particular} sequence (of measure 0).

Martin-L{\"o}f (M.L.) randomness is the standard notion for
randomness of individual sequences \cite{MartinLoef:66,Li:97}. A
M.L.-random sequence passes {\em all} thinkable effective
randomness tests, e.g.\ the law of large numbers, the law of the
iterated logarithm, etc. In particular, the set of all
$\mu$-random sequences has $\mu$-measure 1.
It is natural to ask whether $M_n$ converges to $\mu_n$ (in
difference or ratio) individually for all M.L.-random sequences.
Clearly, Solomonoff's result shows that convergence may at most
fail for a set of sequences with $\mu$-measure zero. A convergence
result for M.L.-random sequences would be particularly interesting and
natural in this context, since M.L.-randomness can be defined in
terms of $M$ itself \cite{Levin:73random}.
Despite several attempts to solve this problem
\cite{Vovk:87,Vitanyi:00,Hutter:03unipriors}, it remained open
\cite{Hutter:03mlconv}.

In this paper we construct an M.L.-random sequence and show the
existence of a universal semimeasure which does not converge on
this sequence, hence answering the open question negatively for
some $M$. It remains open whether there exist (other) universal
semimeasures, probably with particularly interesting additional
structure and properties, for which M.L.-convergence holds. The
main positive contribution of this work is the construction of a
non-universal enumerable semimeasure $W$ which M.L.-converges to
$\mu$ as desired. As an intermediate step we consider the
incomputable measure $\hat D$, defined as a mixture over all
computable measures. We show posterior M.L.-convergence of $W$ to
$\hat D$ and of $\hat D$ to $\mu$. The Hellinger distance
measuring closeness of two posterior distributions plays a central
role in this work.

The paper is organized as follows:
In Section~\ref{secUniM} we give basic notation and results (for
strings, numbers, sets, functions, asymptotics, computability
concepts, prefix Kolmogorov complexity), and define and discuss
the concepts of (universal) (enumerable) (semi)measures.
Section~\ref{secConv} summarizes Solomonoff's and G\'acs' results
on posterior convergence of $M$ to $\mu$ with probability 1. Both
results can be derived from a bound on the expected Hellinger sum.
We present an improved bound on the expected exponentiated
Hellinger sum, which implies very strong assertions on the
convergence rate.
In Section~\ref{secMLNconv} we investigate whether convergence for
all Martin-L{\"o}f random sequences hold. We construct a universal
semimeasure $M$ and an $\mu$-M.L.-random sequence on which $M$
does not converge to $\mu$ for some computable $\mu$.
In Section~\ref{secMLconv} we present our main positive result. We
derive a finite bound on the Hellinger sum between $\mu$ and $\hat
D$, which is exponential in the randomness deficiency of the
sequence and double exponential in the complexity of $\mu$. This
implies that the posterior of $\hat D$ M.L.-converges to $\mu$.
Finally, in Section~\ref{secNUESW} we show that $W$ is
non-universal and asymptotically M.L.-converges to $\hat D$.
Section~\ref{secDisc} contains discussion and outlook.

\section{Notation \& Universal Semimeasures $\MM$}\label{secUniM}

\noindent {\bf Strings.}
Let $i,k,n,t\in\Set N=\{1,2,3,...\}$ be natural numbers,
$x,y,z\in\X^*=\bigcup_{n=0}^\infty\X^n$ be finite strings of
symbols over finite alphabet $\X\ni a,b$.
We denote strings $x$ of length $\l(x)=n$ by
$x=x_1x_2...x_n\in\X^n$ with $x_t\in\X$ and further abbreviate
$x_{k:n}:=x_k x_{k+1}...x_{n-1}x_n$ for $k\leq n$, and $x_{<n}:=x_1...
x_{n-1}$, and $\epstr=x_{<1}=x_{n+1:n}\in\X^0=\{\epstr\}$ for the
empty string. Let $\omega=x_{1:\infty}\in\X^\infty$ be a generic
and $\alpha\in\X^\infty$ a specific infinite sequence. For a
given sequence $x_{1:\infty}$ we say that $x_t$ is on-sequence and
$\bar x_t\neq x_t$ is off-sequence. $x'_t$ may be on- or
off-sequence.
We identify strings with natural numbers (including zero, $\X^*\cong
\Set N\cup\{0\}$).

\noindent {\bf Sets and functions.}
$\Set Q$, $\Set R$, $\Set R_+:=[0,\infty)$ are the sets of fractional,
real, and non-negative real numbers, respectively. $\#\cal S$
denotes the number of elements in set $\cal S$, $\ln()$ the
natural and $\lb()$ the binary logarithm.

\noindent {\bf Asymptotics.}
We abbreviate $\lim_{n\to\infty}[f(n)-g(n)]=0$ by
$f(n)\toinfty{n}g(n)$ and say $f$ converges to $g$, without
implying that $\lim_{n\to\infty}g(n)$ itself exists. We write
$f(x)\leqm g(x)$ for $f(x)=O(g(x))$
and $f(x)\leqa g(x)$ for $f(x)\leq
g(x)+O(1)$.

\noindent {\bf Computability.}
A function $f:\S\to\Set R\cup\{\infty\}$ is said to be enumerable
(or lower semi-computable) if the set $\{(x,y)\,:\,y<f(x),\,
x\in\S,\, y\in\Set Q\}$ is recursively enumerable. $f$ is
co-enumerable (or upper semi-computable) if $[-f]$ is enumerable.
$f$ is computable (or estimable or recursive) if $f$ and $[-f]$ are
enumerable. $f$ is approximable (or limit-computable) if there is
a computable function $g:\S\times\Set N\to\Set R$ with
$\lim_{n\to\infty}g(x,n)=f(x)$.
The set of enumerable functions is recursively enumerable.

\noindent {\bf Complexity.}
The conditional prefix (Kolmogorov) complexity
$K(x|y):=\min\{\l(p):U(y, p)=x \mbox{ halts} \}$ is the length of
the shortest binary program $p\in\B^*$ on a universal prefix
Turing machine $U$ with output $x\in\X^*$ and input $y\in\X^*$
\cite{Li:97}. $K(x):=K(x|\epstr)$.
For non-string objects $o$ we define
$K(o):=K(\langle o\rangle)$, where $\langle o\rangle\in\X^*$ is
some standard code for $o$. In particular, if $(f_i)_{i=1}^n$ is
an enumeration of all enumerable functions, we define
$K(f_i)=K(i)$.
We only need the following elementary properties: %
The co-enumerability of $K$, %
the upper bounds $K(x|\l(x))\leqa\l(x)\lb|\X|$ and $K(n)\leqa 2\lb n$, %
and $K(x|y)\leqa K(x)$, %
subadditivity $K(x)\leqa K(x,y)\leqa K(y)+K(x|y)$, and %
information non-increase $K(f(x))\leqa K(x)+K(f)$ for recursive $f:\X^*\to\X^*$. %

We need the concepts of (universal) (semi)measures for
strings \cite{Zvonkin:70}.

\fdefinition{defSemi}{(Semi)measures}{
We call $\nu:\X^*\to[0,1]$ a semimeasure if
$\nu(x)\geq\sum_{a\in\X}\nu(xa)\,\forall x\in\X^*$, and a
(probability) measure if equality holds and $\nu(\epstr)=1$.
$\nu(x)$ denotes the $\nu$-probability that a sequence starts with
string $x$. Further, $\nu(a|x):={\nu(xa)\over\nu(x)}$ is the
posterior $\nu$-probability that the next symbol is $a\in\X$,
given sequence $x\in\X^*$.
}

\fdefinition{defUniM}{Universal semimeasures $\MM$}{
A semimeasure $\MM$ is called a universal element of
a class of semimeasures $\M$, if
\beqn
  \mbox{ $\MM\in\M$ and
    $\forall\nu\in\M\;\exists w_\nu>0 : \MM(x)\geq w_\nu\!\cdot\!\nu(x)\;\forall x\in\X^*$.}
\eeqn
}
From now on we consider the (in a sense) largest class $\M$ which
is relevant from a constructive point of view (but see
\cite{Schmidhuber:02gtm,Hutter:03unipriors}
for even larger constructive classes), namely the class of {\em
all} semimeasures, which can be enumerated (=effectively be
approximated) from below:
\beq\label{Mclassdef}
  \M:= \;\mbox{class of all enumerable semimeasures}.
\eeq
Solomonoff \cite[Eq.(7)]{Solomonoff:64} defined the universal
posterior $\MM(x|y)=M(xy)/M(y)$ with
$M(x)$ defined as the probability that the output of a
universal monotone Turing machine starts with $x$ when provided
with fair coin flips on the input tape.
Levin \cite{Zvonkin:70} has shown that this $M$ is a universal
enumerable semimeasure.
Another possible definition of $M$ is as a (Bayes) mixture
\cite{Solomonoff:64,Zvonkin:70,Solomonoff:78,Li:97,Hutter:03unipriors}:
$\tilde M(x)=\sum_{\nu\in\M}2^{-K(\nu)}\nu(x)$,
where $K(\nu)$ is the length of the shortest program computing
function $\nu$.
Levin \cite{Zvonkin:70} has shown that the class of {\em all}
enumerable semimeasures is enumerable (with repetitions), hence
$\tilde M$ is enumerable, since $K$ is co-enumerable. Hence
$\tilde\MM\in\M$, which implies
\beq\label{Mdom}
  M(x) \;\geq\; w_{\tilde M} \tilde M(x)
  \;\geq\; w_{\tilde M} 2^{-K(\nu)}\nu(x)
  \;=\; w'_\nu\nu(x),
  \qmbox{where} w'_\nu\eqm 2^{-K(\nu)}.
\eeq
Up to a multiplicative constant, $\MM$ assigns higher probability
to all $x$ than any other enumerable semimeasure.
All $M$ have the same very slowly decreasing (in $\nu$)
domination constants $w'_\nu$, essentially because $\MM\in\M$.
We drop the prime from $w'_\nu$ in the following.
The mixture definition $\tilde M$ immediately generalizes to
arbitrary weighted sums of (semi)measures over other countable
classes than $\M$, but the class may not contain the mixture, and
the domination constants may be rapidly decreasing. We will
exploit this for the construction of the non-universal semimeasure
$W$ in Sections~\ref{secMLconv} and \ref{secNUESW}.

\section{Posterior Convergence with Probability 1}\label{secConv}

The following convergence results for $M$ are well-known
\cite{Solomonoff:78,Li:97,Hutter:02spupper}.

\ftheorem{thConv}{Convergence of $M$ to $\mu$ w.p.1}{
For any universal semimeasure $M$ and any computable measure $\mu$ it holds:
\beqn
 \mbox{$M(x'_n|x_{<n}) \to \mu(x'_n|x_{<n})$
 for any $x'_n$ and
 ${M(x_n|x_{<n})\over\mu(x_n|x_{<n})} \to 1$, both
 w.p.1 for \ifconf$n\!\to\!\infty$\else $n\to\infty$.\fi}
\eeqn
}

The first convergence in difference is Solomonoff's
\cite{Solomonoff:78} celebrated convergence result. The second
convergence in ratio has first been derived by G\'acs
\cite{Li:97}.
Note the subtle difference between the two convergence results.
For {\em any} sequence $x'_{1:\infty}$ (possibly
constant and not necessarily random),
$M(x'_n|x_{<n})-\mu(x'_n|x_{<n})$ converges to zero w.p.1
(referring to $x_{1:\infty}$), but no statement is possible for
$M(x'_n|x_{<n})/\mu(x'_n|x_{<n})$, since
$\lim\,\inf\mu(x'_n|x_{<n})$ could be zero. On the other hand, if
we stay {\em on}-sequence ($x'_{1:\infty} = x_{1:\infty}$), we
have $M(x_n|x_{<n})/\mu(x_n|x_{<n}) \to 1$ (whether
$\inf\mu(x_n|x_{<n})$ tends to zero or not does not matter).
Indeed, it is easy to give an example where
$M(x'_n|x_{<n})/\mu(x'_n|x_{<n})$ diverges. For
$\mu(1|x_{<n})=1-\mu(0|x_{<n})=\odt n^{-3}$ we get
$\mu(0_{1:n})=\prod_{t=1}^n(1-\odt
t^{-3})\stackrel{n\to\infty}\longrightarrow c=0.450...>0$, i.e.\
$0_{1:\infty}$ is $\mu$-random. On the other hand, one can show
that $M(0_{<n})=O(1)$ and $M(0_{<n}1)\eqm 2^{-K(n)}$, which
implies ${M(1|0_{<n})\over \mu(1|0_{<n})} \eqm n^3\cdot
2^{-K(n)}\geqm n \to\infty$ for $n\to\infty$ ($K(n)\leqa
2\log n$).

Theorem~\ref{thConv} follows from (the discussion after) Lemma
\ref{lemHBounds} due to $M(x)\geq w_\mu\mu(x)$. Actually the Lemma
strengthens and generalizes Theorem~\ref{thConv}.
In the following we denote expectations w.r.t.\ measure $\rho$ by
$\E_\rho$, i.e.\ for a function $f:\X^n\to\Set R$,
$\E_\rho[f]=\sum'_{x_{1:n}}\rho(x_{1:n})f(x_{1:n})$, where $\sum'$
sums over all $x_{1:n}$ for which $\rho(x_{1:n})\neq 0$. Using
$\sum'$ instead $\sum$ is important for partial functions $f$
undefined on a set of $\rho$-measure zero. Similarly $\P_{\!\rho}$
denotes the $\rho$-probability.

\flemma{lemHBounds}{Expected Bounds on Hellinger Sum}{
Let $\mu$ be a measure and $\nu$ be a semimeasure with $\nu(x)\geq
w\!\cdot\!\mu(x)$ $\forall x$. Then the following bounds on the
Hellinger distance $h_t(\nu,\mu|\o_{<t}) :=
\sum_{a\in\X}(\sqrt{\nu(a|\o_{<t})}-\sqrt{\mu(a|\o_{<t})}\,)^2$
hold:
\beqn
  \sum_{t=1}^\infty\E{\textstyle\left[\!\left(\sqrt{{\nu(\o_t|\o_{<t})
       \over\mu(\o_t|\o_{<t})}}\!-\!1\right)^2\right]}
  \;\stackrel{(i)}\leq\; \sum_{t=1}^\infty\E[h_t]
  \;\stackrel{(ii)}\leq\; 2\ln\{\E[\exp(\odt\sum_{t=1}^\infty h_t)]\}
  \;\stackrel{(iii)}\leq\; \ln w^{-1}
\eeqn
where $\E$ means expectation w.r.t.\ $\mu$.
}

The $\ln w^{-1}$-bounds on the
first and second expression have first been derived in
\cite{Hutter:02spupper}, the second being a variation of
Solomonoff's bound
$\sum_n\E[(\nu(0|x_{<n})-\mu(0|x_{<n}))^2]\leq\odt\ln w^{-1}$.
If sequence $x_1x_2...$ is sampled from the probability measure
$\mu$, these bounds imply
\beqn
 \mbox{$\nu(x'_n|x_{<n}) \to \mu(x'_n|x_{<n})$
 for any $x'_n$ and
 ${\nu(x_n|x_{<n})\over\mu(x_n|x_{<n})} \to 1$, both
 w.p.1 for $n\to\infty$},
\eeqn
where w.p.1 stands here and in the following for `with
$\mu$-probability 1'.

Convergence is ``fast'' in the following sense: The second bound
($\sum_t\E[h_t]\leq \ln w^{-1}$) implies that the expected number
of times $t$ in which $h_t\geq\eps$ is finite and bounded by
${1\over\eps}\ln w^{-1}$. The new third bound represents a significant
improvement. It implies by means of a Markov inequality
that the probability of even only marginally exceeding this number
is extremely small, and that $\sum_t h_t$ is very unlikely to
exceed $\ln w^{-1}$ by much. More precisely:
\beqn\textstyle
  \P[\#\{t:h_t\geq\eps\}\geq{\textstyle{1\over\eps}}(\ln w^{-1}+c)]
  \;\leq\; \P[\sum_t h_t\geq \ln w^{-1}+c]
\eeqn\vspace{-3ex}
\beqn\textstyle
  \;=\; \P[\exp(\odt\sum_t h_t)\geq\e^{c/2}w^{-1/2}]
  \;\leq\; \sqrt{w}\E[\exp(\odt\sum_t h_t)]\e^{-c/2}
  \;\leq\; \e^{-c/2}.
\eeqn

\paragraph{Proof}
We use the abbreviations $\rho_t=\rho(x_t|x_{<t})$
and $\rho_{1:n}=\rho_1\cdot...\cdot\rho_n=\rho(x_{1:n})$ for
$\rho\in\{\mu,\nu,R,N,...\}$ and
$h_t=\sum_{x_t}(\sqrt{\nu_t}-\sqrt{\mu_t})^2$.

$(i)$ follows from
\beqn
  \E[({\textstyle\sqrt{\nu_t\over\mu_t}}-1)^2|x_{<t}]
  \;\equiv \sum_{x_t:\mu_t\neq 0}\mu_t({\textstyle\sqrt{\nu_t\over\mu_t}}-1)^2
  = \sum_{x_t:\mu_t\neq 0}(\sqrt{\nu_t}-\sqrt{\mu_t})^2
  \;\leq\; h_t
\eeqn
by taking the expectation $\E[]$ and sum $\sum_{t=1}^\infty$.

$(ii)$ follows from Jensen's inequality
$\exp(\E[f])\leq\E[\exp{(f)}]$ for $f=\odt\sum_t h_t$.

$(iii)$ We exploit a construction
used in \cite[Thm.1]{Vovk:87}. For discrete (semi)measures
$p$ and $q$ with $\sum_i p_i=1$ and $\sum_i q_i\leq 1$ it holds:
\beq\label{eqsh}
  \sum_i\sqrt{p_i q_i}
  \;\leq\; 1-\odt\sum_i(\sqrt{p_i}-\sqrt{q_i})^2
  \;\leq\; \exp[-\odt\sum_i(\sqrt{p_i}-\sqrt{q_i})^2].
\eeq
The first inequality is obvious after multiplying out the second
expression. The second inequality follows from $1-x\leq\e^{-x}$.
Vovk \cite{Vovk:87} defined a measure $R_t:=\sqrt{\mu_t \nu_t}/N_t$
with normalization $N_t:=\sum_{x_t}\sqrt{\mu_t \nu_t}$.
Applying (\ref{eqsh}) for measure $\mu$ and semimeasure $\nu$ we get
$N_t\leq\exp(-\odt h_t)$. Together with $\nu(x)\geq w\cdot\mu(x)$
$\forall x$ this implies
\beqn
  \prod_{t=1}^n R_t
  \;=\; \prod_{t=1}^n {\sqrt{\mu_t \nu_t}\over N_t}
  \;=\; {\sqrt{\mu_{1:n}\nu_{1:n}}\over N_{1:n}}
  \;=\; \mu_{1:n} {\sqrt{\nu_{1:n}\over\mu_{1:n}}}N_{1:n}^{-1}
  \;\geq\; \mu_{1:n} \sqrt{w} \exp(\odt\sum_{t=1}^n h_t).
\eeqn
Summing over $x_{1:n}$ and exploiting $\sum_{x_t}R_t=1$ we get
$1\geq \sqrt{w}\E[\exp(\odt\sum_t h_t)$], which proves $(iii)$.

The bound and proof may be generalized to
$1\geq w^\kappa\E[\exp(\odt\sum_t\sum_{x_t}(\nu_t^\kappa-\mu_t^\kappa)^{1/\kappa})]$ with
$0\leq\kappa\leq\odt$ by defining
$R_t=\mu_t^{1-\kappa}\nu_t^\kappa/N_t$ with
$N_t=\sum_{x_t}\mu_t^{1-\kappa}\nu_t^\kappa$ and exploiting $\sum_i
p_i^{1-\kappa}q_i^\kappa \leq
\exp(-\odt\sum_i(p_i^\kappa-q_i^\kappa)^{1/\kappa})$.
\qed

One can show that the constant $\odt$ in Lemma \ref{lemHBounds}
can essentially not been improved. Increasing it to a constant
$\alpha>1$ makes the expression infinite for some (Bernoulli)
distribution $\mu$ (however we choose $\nu$). For $\nu=M$ the
expression can become already infinite for $\alpha>\odt$ and some
computable measure $\mu$.

\section{Non-Convergence in Martin-L{\"o}f Sense}\label{secMLNconv}

Convergence of $M(x_n|x_{<n})$ to $\mu(x_n|x_{<n})$ with
$\mu$-probability 1 tells us that $M(x_n|x_{<n})$ is close to
$\mu(x_n|x_{<n})$ for sufficiently large $n$ on ``most''
sequences $x_{1:\infty}$. It says nothing whether convergence is
true for any {\em particular} sequence (of measure 0).
Martin-L\"{o}f randomness can be used to capture convergence
properties for individual sequences.
Martin-L\"{o}f randomness is a very important concept of
randomness of individual sequences, which is closely related to
Kolmogorov complexity and Solomonoff's universal semimeasure
$M$. Levin gave a characterization equivalent to Martin-L\"{o}f's
original definition \cite{Levin:73random}:

\fdefinition{defML}{Martin-L\"{o}f random sequences}{
A sequence $\o=\o_{1:\infty}$ is $\mu$-Martin-L\"{o}f random
($\mu$.M.L.) iff there is a constant $c<\infty$ such that
$\MM(\o_{1:n})\leq c\cdot \mu(\o_{1:n})$ for all $n$.
Moreover, $d_\mu(\o):=\sup_n\{\lb {\MM(\o_{1:n})\over\mu(\o_{1:n})}\}\leq
\lb c$ is called the randomness deficiency of $\o$.
}
One can show that an M.L.-random sequence $x_{1:\infty}$ passes
{\em all} thinkable effective randomness tests, e.g.\ the law of
large numbers, the law of the iterated logarithm, etc. In
particular, the set of all $\mu$.M.L.-random sequences has
$\mu$-measure 1.

The open question we study in this section is whether $M$ converges
to $\mu$ (in difference or ratio) individually for all
Martin-L\"{o}f random sequences. Clearly, Theorem~\ref{thConv}
implies that convergence $\mu$.M.L. may at most fail for a set of
sequences with $\mu$-measure zero. A convergence M.L.\ result
would be particularly interesting and natural for $M$, since
M.L.-randomness can be defined in terms of $\MM$ itself (Definition
\ref{defML}).

The state of the art regarding this problem may be summarized as
follows: \cite{Vovk:87} contains a (non-improvable?) result which
is slightly too weak to imply M.L.-convergence,
\cite[Thm.5.2.2]{Li:97} and \cite[Thm.10]{Vitanyi:00} contain an
erroneous proof for M.L.-convergence, and
\cite{Hutter:03unipriors} proves a theorem indicating that the
answer may be hard and subtle (see \cite{Hutter:03unipriors} for
details).

The main contribution of this section is a partial answer to this
question. We show that M.L.-convergence fails at least for some
universal semimeasures:

\ftheorem{thMnonConv}{Universal semimeasure non-convergence}{
There exists a universal semimeasure $M$ and a computable measure
$\mu$ and a $\mu$.M.L.-random sequence $\alpha$, such that
\vspace{-2ex}\beqn
  M(\a_n|\a_{<n}) \;\;\not\!\!\!\longrightarrow \mu(\a_n|\a_{<n})
  \qmbox{for} n\to\infty.
\eeqn
}
This implies that also $M_n/\mu_n$ does not converge (since $\mu_n\leq
1$ is bounded). We do not know whether Theorem~\ref{thMnonConv}
holds for {\em all} universal semimeasures.
The proof idea is to construct an enumerable (semi)measure $\nu$
such that $\nu$ dominates $M$ on some $\mu$-random sequence
$\alpha$, but $\nu(\a_n|\a_{<n})\not\to\mu(\a_n|\a_{<n})$. Then we mix $M$ to
$\nu$ to make $\nu$ universal, but with larger contribution from
$\nu$, in order to preserve non-convergence. There is also
non-constructive proof showing that an arbitrary small
contamination with $\nu$ can lead to non-convergence. We only
present the constructive proof.

\paragraph{Proof}
We consider binary alphabet $\X=\{0,1\}$ only. Let
$\mu(x)=\lambda(x):=2^{-\l(x)}$ be the uniform measure.
We define the sequence $\a$ as the (in a sense)
lexicographically first (or equivalently left-most in the tree of
sequences) $\lambda$.M.L.-random sequence. Formally we define
$\a$, inductively in $n=1,2,3,...$ by
\beq\label{defalpha}
 \mbox{$\a_n=0$ if $M(\a_{<n}0)\leq 2^{-n}$, and $\a_n=1$ else.}
\eeq
We know that $M(\epstr)\leq 1$ and $M(\a_{<n}0)\leq 2^{-n}$ if
$\a_n=0$. Inductively, assuming $M(\a_{<n})\leq 2^{-n+1}$ for
$\a_n=1$ we have $2^{-n+1}\geq M(\a_{<n})\geq
M(\a_{<n}0)+M(\a_{<n}1) \geq 2^{-n}+M(\a_{<n}1)$ since $M$ is a
semimeasure, hence $M(\a_{<n}1)\leq 2^{-n}$.
Hence
\beq\label{arand}
  \mbox{$M(\a_{1:n})\leq 2^{-n}\equiv\lambda(\a_{1:n})\,\forall n$,
  i.e.\ $\a$ is $\lambda$.M.L.-random.}
\eeq%
Let $M^t$ with $t=1,2,3,...$ be computable approximations of $M$,
which enumerate $M$, i.e.\ $M^t(x)\nearrow M(x)$ for $t\to\infty$.
W define $\alpha^t$ like $\a$ but with $M$ replaced by $M^t$ in
the definition. $M^t\nearrow M$ implies $\a^t\!\!\nearrow\a$
(lexicographically increasing).
We define an enumerable semimeasure $\nu$ as follows:
\beq\label{nudef}
   \nu^t(x):=\left\{
   \begin{array}{ccl}
     2^{-t} & \mbox{if} & \l(x)=t \qmbox{and} x<\a_{1:t}^t \\
     0      & \mbox{if} & \l(x)=t \qmbox{and} x\geq\a_{1:t}^t \\
     0      & \mbox{if} & \l(x)>t \\
     \nu^t(x0)\!+\!\nu^t(x1) & \mbox{if} & \l(x)<t \\
   \end{array}\right.
\eeq
where $<$ is the lexicographical ordering on sequences. $\nu^t$ is a
semimeasure, and with $\alpha^t$ also $\nu^t$ is computable and
monotone increasing in $t$, hence $\nu:=\lim_{t\to\infty}\nu^t$ is
an enumerable semimeasure (indeed, ${\nu(x)\over\nu(\epstr)}$ is a
measure). We could have defined a $\nu_{tn}$ by replacing
$\alpha_{1:t}^t$ with $\alpha_{1:t}^n$ in (\ref{nudef}). Since
$\nu_{tn}$ is monotone increasing in $t$ and $n$, any order of
$t,n\to\infty$ leads to $\nu$, so we have chosen arbitrarily $t=n$.
By induction (starting from $\l(x)=t$) it follows that
\beqn
  \nu^t(x)=2^{-\l(x)} \qmbox{if} x<\a_{1:\l(x)}^t \qmbox{and} \l(x)\leq t,
  \qquad\qquad
  \nu^t(x)=0 \qmbox{if} x>\a_{1:\l(x)}^t
\eeqn
On-sequence, i.e.\ for $x=\a_{1:n}$, $\nu^t$ is somewhere
in-between $0$ and $2^{-\l(x)}$. Since sequence $\a:=\lim_t\a^t$
is $\lambda$.M.L.-random it contains $01$ infinitely often,
actually $\a_n\a_{n+1}=01$ for a non-vanishing fraction of $n$. In
the following we fix such an $n$. For $t\geq n$ we get
\beqn
  \nu^t(\a_{<n})
  = \nu^t(\a_{<n}0) \!+\! \nu^t(\underbrace{\a_{<n}1}_{\nq\nq\nq>\a_{1:n}\geq\a_{1:n}^t,\text{ since }\a_n=0\nq\nq\nq})
  = \nu^t(\a_{<n}0)
  = \nu^t(\a_{1:n})
  \quad\Rightarrow\quad \nu(\a_{<n})=\nu(\a_{1:n})
\eeqn
This ensures $\nu(\a_n|\a_{<n})=1\neq\odt=\lambda_n$. For
$t>n$ large enough such that $\a_{1:n+1}^t=\a_{1:n+1}$ we get:
\beqn
  \nu^t(\a_{1:n})
  = \nu^t(\a_{1:n}^t)
  \geq \nu^t(\underbrace{\a_{1:n}^t 0}_{\nq\nq\nq<\a_{1:n+1}^t,\text{ since }\a_{n+1}=1\nq\nq\nq})
  = 2^{-n-1}
  \quad\Rightarrow\quad \nu(\a_{1:n})\geq 2^{-n-1}
\eeqn
This ensures $\nu(\a_{1:n})\geq 2^{-n-1}\geq\odt M(\a_{1:n})$ by
(\ref{arand}). Let $M$ be any universal semimeasure and
$0<\gamma<{1\over 5}$. Then $M'(x):=(1-\gamma)\nu(x)+\gamma
M(x)\,\forall x$ is also a universal semimeasure with
\bqan
  M'(\a_n|\a_{<n})
   & & \ifconf\else\nq\fi=\;\; {(1\!-\!\gamma)\nu(\a_{1:n})+\gamma M(\a_{1:n})\over (1\!-\!\gamma)\nu(\a_{<n})+\gamma M(\a_{<n})}
  \;\mathop{\rule{0ex}{2.5ex}\geq}^{\displaystyle\mathop{\rule{0ex}{2.5ex}
    \downarrow}^{\makebox[0ex]{\footnotesize
    $M(\a_{<n})\leq 2^{-n+1}$ and $M(\a_{1:n})\geq 0$}}}\;
  {(1\!-\!\gamma)\nu(\a_{1:n})\over (1\!-\!\gamma)\nu(\a_{<n})+\gamma 2^{-n+1}}
\\
   & & \ifconf\else\nq\fi\mathop=_{\displaystyle\mathop{\rule{0ex}{3ex}
    \uparrow}_{\rule{0ex}{2ex}\makebox[0ex]{\footnotesize
    $\nu(\a_{<n})=\nu(\a_{1:n})$}}}\;\;
  {1\!-\!\gamma\over 1\!-\!\gamma + \gamma 2^{-n+1}/\nu(\a_{1:n})}
  \;\;\mathop\geq_{\displaystyle\mathop{\rule{0ex}{3ex}
    \uparrow}_{\rule{0ex}{2ex}\makebox[0ex]{\footnotesize
    $\nu(\a_{1:n})\geq 2^{-n-1}$}}}\;\;
  {1\!-\!\gamma\over 1+3\gamma}
  \;\;>\;\; {1\over 2}.
\eqan
For instance for $\gamma={1\over 9}$ we have
$M'(\a_n|\a_{<n})\geq{2\over 3}\neq\odt=\lambda(\a_n|\a_{<n})$ for
a non-vanishing fraction of $n$'s. \qed

A converse of Theorem~\ref{thMnonConv} can also be shown:

\ftheorem{thConvNR}{Convergence on non-random sequences}{
For every universal semimeasure $M$ there exist computable
measures $\mu$ and non-$\mu$.M.L.-random sequences $\a$ for which
$M(\a_n|\a_{<n})/\mu(\a_n|\a_{<n})\to 1$.
}

\section{Convergence in Martin-L{\"o}f Sense}\label{secMLconv}

In this and the next section we give a positive answer to the
question of posterior M.L.-convergence to $\mu$. We consider
general finite alphabet $\X$.

\ftheorem{thMLconv}{Universal predictor for M.L.-random sequences}{
There exists an enumerable semimeasure $W$ such that for every
computable measure $\mu$ and every $\mu$.M.L.-random sequence $\omega$,
the posteriors converge to each other: 
\beqn\textstyle
  W(a|\o_{<t})\toinfty{t}\mu(a|\o_{<t}) \qmbox{for all} a\in\X \qmbox{if}
  d_\mu(\o)<\infty.
\eeqn
}

The semimeasure $W$ we will construct is not universal in the
sense of dominating all enumerable semimeasures, unlike $M$.
Normalizing $W$ shows that there is also a measure whose posterior
converges to $\mu$, but this measure is not enumerable, only
approximable. For proving Theorem~\ref{thMLconv} we first define
an intermediate measure $D$ as a mixture over all computable
measures, which is not even approximable. Based on Lemmas
\ref{lemHBounds},\ref{lemHChain},\ref{lemE2I},
Proposition~\ref{proDtomu} shows that $D$ M.L.-converges to $\mu$.
We then define the concept of quasimeasures and an enumerable
semimeasure $W$ as a mixture over all enumerable quasimeasures.
Proposition~\ref{proWtoD} shows that $W$ M.L.-converges to $D$.
Theorem~\ref{thMLconv} immediately follows from
Propositions~\ref{proDtomu} and~\ref{proWtoD}.

\flemma{lemHChain}{Hellinger Chain}{
Let $h(\v p,\v q):=\sum_{i=1}^N(\sqrt{p_i}-\sqrt{q_i})^2$ be the
Hellinger distance between $\v p=(p_i)_{i=1}^N\in\Set R_+^N$ and
$\v q=(q_i)_{i=1}^N \in\Set R_+^N$. Then
\beqn
\begin{array}{rlcll}
  i)   & \mbox{for}\; \v p,\v q,\v r\in\Set R_+^N     & h(\v p,\v q)
       & \leq & (1+\beta)\,h(\v p,\v r)+(1+\beta^{-1})\,h(\v r,\v q), \;\mbox{any}\; \beta>0
\\
  ii)  & \mbox{for}\; \v p^1,...,\v p^m \in\Set R_+^N & h(\v p^1,\v p^m)
       & \leq & \displaystyle 3\sum_{k=2}^m k^2\,h(\v p^{k-1},\v p^k)
\end{array}
\eeqn
}

\paragraph{Proof} $(i)$
For any $x,y\in\Set R$ and $\beta>0$ we have $(x+y)^2\leq
(1+\beta)x^2+(1+\beta^{-1})y^2$. Inserting
$x=\sqrt{p_i}-\sqrt{r_i}$ and $y=\sqrt{r_i}-\sqrt{q_i}$ and
summing over $i$ proves $(i)$.

$(ii)$ Apply $(i)$ for the triples $(\v p^k,\v p^{k+1},\v p^m)$
for and in order of $k=1,2,...,m-2$ with $\beta=\beta_k=k(k+1)$ and
finally use $\prod_{j=1}^{k-2}(1+\beta_j^{-1})\leq \e \leq 3$. \qed

We need a way to convert expected bounds to bounds on individual
M.L.\ random sequences, sort of a converse of ``M.L.\ implies
w.p.1''. Consider for instance the Hellinger sum
$H(\o):=\sum_{t=1}^\infty h_t(\mu,\rho)/\ln w^{-1}$ between two
computable measures $\rho\geq w\!\cdot\!\mu$. Then $H$ is an
enumerable function and Lemma~\ref{lemHBounds} implies $\E[H]\leq
1$, hence $H$ is an integral $\mu$-test. $H$ can be increased to an enumerable
$\mu$-submartingale $\bar H$. The universal $\mu$-submartingale
$M/\mu$ multiplicatively dominates all enumerable submartingales
(and hence $\bar H$).
Since $M/\mu\leq 2^{d_\mu(\o)}$, this implies the desired bound
$H(\o) \leqm 2^{d_\mu(\o)}$ for individual $\o$.
We give a self-contained direct proof, explicating all important
constants.

\flemma{lemE2I}{Expected to Individual Bound}{
Let $F(\o)\geq 0$ be an enumerable function and $\mu$ be an
enumerable measure and $\eps>0$ be co-enumerable. Then:
\beqn
  \qmbox{If} \E_\mu[F] \;\leq\; \eps
  \qmbox{then} F(\o) \;\leqm\; \eps\!\cdot\! 2^{K(\mu,F,\,^1\!/\eps)+d_\mu(\o)}
  \quad \forall\o
\eeqn
where $d_\mu(\o)$ is the $\mu$-randomness deficiency of $\o$ and
$K(\mu,F,\,^1\!/\eps)$ is the length of the shortest program for $\mu$,
$F$, and $^1\!/\eps$.
}

Lemma~\ref{lemE2I} roughly says that for $\mu$, $F$, and
$\eps\eqm\E_\mu[F]$ with short program ($K(\mu,F,^1\!/\eps)=O(1)$) and
$\mu$-random $\o$ ($d_\mu(\o)=O(1)$) we have $F(\o)\leqm
\E_\mu[F]$.

\paragraph{Proof}
Let $F(\o)=\lim_{n\to\infty}F_n(\o)=\sup_n F_n(\o)$ be enumerated
by an increasing sequence of computable functions $F_n(\o)$.
$F_n(\o)$ can be chosen to depend on $\o_{1:n}$ only, i.e.\
$F_n(\o)=F_n(\o_{1:n})$ is independent of $\o_{n+1:\infty}$. Let
$\eps_n\!\!\searrow\eps$ co-enumerate $\eps$. We define
\beqn
  \bar\mu_n(\o_{1:k}) \;:=\; \eps_n^{-1} \nq\ifconf\nq\;\fi \sum_{\o_{k+1:n}\in\X^{n-k}}\nq
  \mu(\o_{1:n})F_n(\o_{1:n}) \;\;\mbox{for}\;\; k\leq n,
  \qmbox{and} \bar\mu_n(\o_{1:k})=0 \;\;\mbox{for}\;\; k>n.
\eeqn
$\bar\mu_n$ is a computable semimeasure for each $n$ (due to
$\E_\mu[F_n]\leq\eps$) and increasing in $n$, since
\bqan
  \bar\mu_n(\o_{1:k}) & \!\!\geq\!\! & 0 \;=\; \bar\mu_{n-1}(\o_{1:k}) \qmbox{for} k\geq n \qmbox{and}
\\
  \bar\mu_n(\o_{<n}) &
  \mathop\geq_{\displaystyle\mathop{\rule{0ex}{3.5ex}\nq
    \uparrow}_{\rule{0ex}{2ex}\makebox[0ex]{\footnotesize
      $\quad F_n\geq F_{n-1}$}}} &
  \!\!\sum_{\o_n\in\X} \eps_n^{-1}\mu(\o_{1:n})F_{n-1}(\o_{<n})
  \;\mathop=_{\displaystyle\mathop{\rule{0ex}{2.5ex}
    \uparrow}_{\rule{0ex}{2ex}\makebox[0ex]{\footnotesize $\mu$ measure}}}\;
  \eps_n^{-1}\mu(\o_{<n})F_{n-1}(\o_{<n})
  \;\mathop\geq_{\displaystyle\mathop{\rule{0ex}{2ex}
    \uparrow}_{\rule{0ex}{2ex}\makebox[0ex]{\footnotesize
      $\quad\eps_n\leq \eps_{n-1}$}}} \;
  \bar\mu_{n-1}(\o_{<n})
\eqan
and similarly for $k<n-1$. Hence $\bar\mu:=\bar\mu_\infty$ is an
enumerable semimeasure (indeed $\bar\mu$ is proportional to a
measure). From dominance (\ref{Mdom}) we get
\beq\label{eqMmuF}
  M(\o_{1:n})
  \;\geqm\; 2^{-K(\bar\mu)}\bar\mu(\o_{1:n})
  \;\geq\; 2^{-K(\bar\mu)}\bar\mu_n(\o_{1:n})
  \;=\; 2^{-K(\bar\mu)} \eps_n^{-1}\mu(\o_{1:n})F_n(\o_{1:n}).
\eeq
In order to enumerate $\bar\mu$, we need to enumerate $\mu$, $F$,
and $\eps^{-1}$, hence $K(\bar\mu)\leqa K(\mu,F,\,^1\!/\eps)$, so
we get
\beqn
  F_n(\o)\;\equiv\; F_n(\o_{1:n})
  \;\leqm\; \eps_n\!\cdot\! 2^{K(\mu,F,^1\!/\eps)}\!\cdot\!\textstyle{M(\o_{1:n})\over\mu(\o_{1:n})}
  \;\leq\; \eps_n\!\cdot\! 2^{K(\mu,F,^1\!/\eps)+d_\mu(\o)}.
\eeqn
Taking the limit $F_n\nearrow F$ and $\eps_n\!\!\searrow\eps$
completes the proof.
\qed

Let $\M=\{\nu_1,\nu_2,...\}$ be an enumeration of all enumerable
semimeasures, $J_k:=\{i\leq k \,:\, \nu_i $ is measure$\}$, and
$\delta_k(x):=\sum_{i\in J_k} \eps_i\nu_i(x)$. The weights
$\eps_i$ need to be computable and exponentially decreasing in $i$
and $\sum_{i=1}^\infty \eps_i\leq 1$. We choose $\eps_i=i^{-6}
2^{-i}$. Note the subtle and important fact that although the
definition of $J_k$ is non-constructive, as a finite set of finite
objects, $J_k$ is decidable (the program is unknowable for large
$k$). Hence, $\delta_k$ is computable, since enumerable measures
are computable.
\beqn
  D(x) \;=\; \delta_\infty(x)
  \;=\; \sum_{i\in J_\infty} \eps_i\nu_i(x)=
  \mbox{mixture of all computable measures}.
\eeqn
In contrast to $J_k$ and $\delta_k$, the set $J_\infty$ and hence
$D$ are neither enumerable nor co-enumerable. We also define the
measures $\hat\delta_k(x):=\delta_k(x)/\delta_k(\epstr)$ and $\hat
D(x):=D(x)/D(\epstr)$. The following Proposition implies posterior
convergence of $D$ to $\mu$ on $\mu$-random sequences.

\fproposition{proDtomu}{Convergence of incomputable measure $\hat D$}{
Let $\mu$ be a computable measure with index $k_0$, i.e.\
$\mu=\nu_{k_0}$. Then for the incomputable measure $\hat D$ and
the computable but non-constructive measures $\hat\delta_{k_0}$ defined above,
the following holds:
\beqn
\begin{array}{rccl}
  i)  & \sum_{t=1}^\infty h_t(\hat\delta_{k_0},\mu)    & \leqa & 2\ln 2 \!\cdot\! d_\mu(\o) +3k_0 \\[1ex]
  ii)   & \sum_{t=1}^\infty h_t(\hat\delta_{k_0},\hat D) & \leqm & k_0^7 2^{k_0+d_\mu(\o)} \\
\end{array}
\eeqn
}

Combining $(i)$ and $(ii)$, using Lemma~\ref{lemHChain}$(i)$, we
get $\sum_{t=1}^\infty h_t(\mu,\hat D) \leq c_\o f(k_0) < \infty$
for $\mu$-random $\o$, which implies $D(b|\o_{<t})\equiv\hat
D(b|\o_{<t}) \to \mu(b|\o_{<t})$. We do not know whether
on-sequence convergence of the ratio holds. Similar bounds hold
for $\hat\delta_{k_1}$ instead $\hat\delta_{k_0}$, $k_1\geq k_0$.
The principle proof idea is to convert the expected bounds of
Lemma~\ref{lemHBounds} to individual bounds, using Lemma
\ref{lemE2I}. The problem is that $\hat D$ is not computable,
which we circumvent by joining with Lemma~\ref{lemHChain}, bounds
on $\sum_th_t(\hat\delta_{k-1},\hat\delta_k)$ for
$k=k_0,k_0+1,...$.

\paragraph{Proof}
$(i)$ Let $H(\o):=\sum_{t=1}^\infty h_t(\hat\delta_{k_0},\mu)$.
$\mu$ and $\hat\delta_{k_0}$ are measures with
$\hat\delta_{k_0}\geq \delta_{k_0}\geq \eps_{k_0}\mu$, since
$\delta_k(\epstr)\leq 1$, $\mu=\nu_{k_0}$ and $k_0\in J_{k_0}$.
Hence, Lemma~\ref{lemHBounds} applies and shows $\E_\mu[\exp(\odt
H)]\leq \eps_{k_0}^{-1/2}$.
$H$ is well-defined and enumerable for $d_\mu(\o)<\infty$, since
$d_\mu(\o)<\infty$ implies $\mu(\o_{1:t})\neq 0$ implies
$\hat\delta_{k_0}(\o_{1:t})\neq 0$. So $\mu(b|\o_{1:t})$ and
$\hat\delta_{k_0}(b|\o_{1:t})$ are well defined and computable
(given $J_{k_0}$). Hence $h_t(\hat\delta_{k_0},\mu)$ is
computable, hence $H(\o)$ is enumerable.
Lemma~\ref{lemE2I} then implies $\exp(\odt H(\o))\leqm
\eps_{k_0}^{-1/2}\cdot 2^{K(\mu,H,\sqrt{\eps}_{k_0})+d_\mu(\o)}$.
We bound
\beqn
  K(\mu,H,\sqrt{\eps}_{k_0})
  \;\leqa\; K(H|\mu,k_0) + K(k_0)
  \;\leqa\; K(J_{k_0}|k_0) + K(k_0)
  \;\leqa\; k_0 + 2\lb k_0.
\eeqn
The first inequality holds, since $k_0$ is the index and hence a
description of $\mu$, and $\eps_*$ is a simple computable
function. $H$ can be computed from $\mu$, $k_0$ and $J_{k_0}$,
which implies the second inequality. The last inequality follows
from $K(k_0)\leqa 2\lb k_0$ and the fact that for each $i\leq k_0$
one bit suffices to specify (non)membership to $J_{k_0}$, i.e.\
$K(J_{k_0}|k_0)\leqa k_0$. Putting everything together we get
\beqn
  H(\o) \;\leqa\; \ln\eps_{k_0}^{-1} + [k_0 + 2\lb k_0 + d_\mu(\o)]2\ln 2
  \;\leqa\; (2\ln 2) d_\mu(\o) + 3k_0.
\eeqn

$(ii)$ Let $H^k(\o):=\sum_{t=1}^\infty h_t(\hat\delta_k,\hat\delta_{k-1})$
and $k>k_0$. $\delta_{k-1}\leq\delta_k$ implies
\beqn
  {\hat\delta_{k-1}(x)\over\hat\delta_k(x)}
  \;\leq\; {\delta_k(\epstr)\over\delta_{k-1}(\epstr)}
  \;\leq\; {\delta_{k-1}(\epstr)+\eps_k\over\delta_{k-1}(\epstr)}
  \;=\; 1+{\eps_k\over\delta_{k-1}(\epstr)}
  \;\leq\; 1+{\eps_k\over\eps_O},
\eeqn
where $O:=\min\{i\in J_{k-1}\}=O(1)$. Note that $J_{k-1}\ni k_0$ is
not empty. Since $\hat\delta_{k-1}$ and $\hat\delta_k$ are
measures, Lemma~\ref{lemHBounds} applies and shows
$\E_{\hat\delta_{k-1}}[H^k]\leq
\ln(1+{\eps_k\over\eps_O})\leq {\eps_k\over\eps_O}$.
Exploiting $\eps_{k_0}\mu\leq\hat\delta_{k-1}$, this implies
$\E_\mu[H^k]\leq {\eps_k\over\eps_O\eps_{k_0}}$. Lemma
\ref{lemE2I} then implies $H^k(\o)\leqm
{\eps_k\over\eps_O\eps_{k_0}}\cdot
2^{K(\mu,H^k,\eps_O\eps_{k_0}/\eps_k)+d_\mu(\o)}$.
Similarly as in $(i)$ we can bound
\beqn
  K(\mu,H^k,\eps_{k_0}/\eps_O\eps_k) \leqa K(J_k|k)+K(k)+K(k_0) \leqa
  k+2\lb k+2\lb k_0, \qmbox{hence}
\eeqn\vspace{-3ex}
\beqn\textstyle
  H^k(\o) \;\leqm\; {\eps_k\over\eps_O\eps_{k_0}}\!\cdot\!k_0^2 k^2 2^k c_\o
  \;\eqm\; k_0^8 2^{k_0}k^{-4}c_\o, \qmbox{where} c_\o:=2^{d_\mu(\o)}.
\eeqn
Chaining this bound via Lemma~\ref{lemHChain}$(ii)$ we get for $k_1>k_0$:
\bqan
  \sum_{t=1}^n h_t(\hat\delta_{k_0},\hat\delta_{k_1})
   &\leq& \sum_{t=1}^n 3 \!\!\sum_{k=k_0+1}^{k_1}\!\! (k\!-\!k_0\!+\!1)^2 h_t(\hat\delta_{k-1},\hat\delta_k)
\\
  &\leq& 3 \!\!\sum_{k=k_0+1}^{k_1}\!\! k^2 H^k(\o)
  \;\leqm\; 3k_0^8 2^{k_0}c_\o \!\!\sum_{k=k_0+1}^{k_1}\!\! k^{-2}
  \;\leq\; 3k_0^7 2^{k_0}c_\o
\eqan
If we now take $k_1\to\infty$ we get $\sum_{t=1}^n
h_t(\hat\delta_{k_0},\hat D) \leqm 3k_0^7
2^{k_0+d_\mu(\o)}$. Finally let $n\to\infty$. \qed

The main properties allowing for proving $\hat D\to\mu$ were that
$\hat D$ is a measure with approximations $\hat\delta_k$, which
are computable in a certain sense. $\hat D$ is a mixture over all
enumerable/computable measures and hence incomputable.

\section{M.L.-Converging Enumerable Semimeasure $W$}\label{secNUESW}

The next step is to enlarge the class of computable measures to an
enumerable class of semimeasures, which are still sufficiently
close to measures in order not to spoil the convergence result.
For convergence w.p.1.\ we could include {\em all} semimeasures
(Theorem~\ref{thConv}). M.L.-convergence seems to require a more
restricted class. Included non-measures need to be zero on long
strings.
We convert semimeasures $\nu$ to ``quasimeasures'' $\tilde\nu$ as
follows:
\beqn
  \tilde\nu(x_{1:n}):=\nu(x_{1:n})
  \qmbox{if} \sum_{y_{1:n}}\nu(y_{1:n}) \;>\; 1- {1\over n}
  \qmbox{and} \nu(x_{1:n}):=0 \qmbox{else.}
\eeqn
If the condition is violated for some $n$ it is also violated for
all larger $n$, hence with $\nu$ also $\tilde\nu$ is a
semimeasure. $\tilde\nu$ is enumerable if $\nu$ is enumerable. So
if $\nu_1,\nu_2,...$ is an enumeration of all enumerable
semimeasures, then $\tilde\nu_1,\tilde\nu_2,...$ is an enumeration
of all enumerable quasimeasures. The for us important properties
are that $\tilde\nu_i\leq\nu_i$ -and- if $\nu_i$ is a measure,
then $\tilde\nu_i\equiv\nu_i$, else $\nu_i(x)=0$ for sufficiently
long $x$.
We define the enumerable semimeasure
\beqn
  W(x):=\sum_{i=1}^\infty\eps_i\tilde\nu_i(x),
  \qmbox{and\ifconf\else note that\fi}
  D(x)=\sum_{i\in J} \eps_i\tilde\nu_i(x)
  \;\;\mbox{with}\;\;
  J:=\{i:\tilde\nu_i\mbox{ is measure}\}
\eeqn
with $\eps_i=i^{-6}2^{-i}$ as before.

\fproposition{proWtoD}{Convergence of enumerable $W$ to incomputable $D$}{
For every computable measure $\mu$ and for $\o$ being
$\mu$-random, the following holds for $t\to\infty$:
\beqn
  (i)   \;\; {W(\o_{1:t})\over D(\o_{1:t})}\to 1, \ifconf\quad\else\qquad\fi
  (ii)  \;\; {W(\o_t|\o_{<t})\over D(\o_t|\o_{<t})} \to 1, \ifconf\quad\else\qquad\fi
  (iii) \;\; W(a|\o_{<t})\to D(a|\o_{<t}) \;\;\forall a\in\X.
\eeqn
}

The intuitive reason for the convergence is that the additional
contributions of non-measures to $W$ absent in $D$ are zero
for long sequences.

\paragraph{Proof} $(i)$\vspace{-3ex}
\beq\label{eqWtoD}
  D(x) \;\leq\; W(x)
  \;=\; D(x) + \sum_{i\not\in J}\eps_i\tilde\nu_i(x)
  \;\leq\; D(x) + \sum_{i=k_x}^\infty\eps_i\tilde\nu_i(x),
\eeq
where $k_x:=\min_i\{i\not\in J:\tilde\nu_i(x)\neq 0\}$. For
$i\not\in J$, $\tilde\nu_i$ is not a measure. Hence
$\tilde\nu_i(x)=0$ for sufficiently long $x$. This implies
$k_x\to\infty$ for $\l(x)\to\infty$, hence $W(x)\to D(x)$ $\forall
x$. To get convergence in ratio we have to assume that
$x=\o_{1:n}$ with $\o$ being $\mu$-random, i.e.\
$c_\o:=\sup_n{M(\o_{1:n})\over\mu(\o_{1:n})}=2^{d_\mu(\o)}<\infty$.
\beqn
  \Rightarrow\; \tilde\nu_i(x)
  \;\leq\; \nu_i(x)
  \;\leq\; {1\over w_{\nu_i}}M(x)
  \;\leq\; {c_\omega\over w_{\nu_i}}\mu(x)
  \;\leq\; {c_\omega\over w_{\nu_i}\eps_{k_0}}D(x),
\eeqn
The last inequality holds, since $\mu$ is a computable measure of
index $k_0$, i.e.\ $\mu=\nu_{k_0}=\tilde\nu_{k_0}$. Inserting
$1/w_{\nu_i}\leq c'\cdot i^2$ for some $c=O(1)$ and
$\eps_i$ we get $\eps_i\tilde\nu_i(x) \leq {c'
c_\o\over\eps_{k_0}}i^{-4}2^{-i}D(x)$, which implies
$\sum_{i=k_x}^\infty \eps_i\tilde\nu_i(x) \leq \eps_x'D(x)$ with
$\eps_x':={2c'c_\o\over\eps_{k_0}}k_x^{-4}2^{-k_x} \to 0$ for
$\l(x)\to\infty$.
Inserting this into (\ref{eqWtoD}) we get
\beqn
  1 \;\leq\; {W(x)\over D(x)} \;\leq\; 1+\eps'_x \;\;\toinfty{\l(x)}\;\; 1
  \qmbox{for $\mu$-random $x$.}
\eeqn

$(ii)$ Obvious from $(i)$ by taking a double ratio.

$(iii)$
Let $a\in\X$. From $W(xa)\geq D(xa)$ ($W\geq D$) and
$W(x)\leq (1+\eps'_x)D(x)$ $(i)$ we get
\bqan
  W(a|x) &\geq& (1+\eps'_x)^{-1} D(a|x) \;\geq\; (1-\eps'_x) D(a|x)
  \qquad \forall a\in\X,
  \qmbox{and} \\
  1-W(a|x) &\geq& \sum_{b\neq a} W(b|x)
  \;\geq\; (1-\eps'_x) \sum_{b\neq a} D(b|x)
  \;=\; (1-\eps'_x) (1-D(a|x)),
\eqan
where we used in the second line that $W$ is a semimeasure and $D$
proportional to a measure. Together this implies
$|W(a|x)-D(a|x)|\leq\eps'_x$.
Since $\eps'_x\to 0$ for $\mu$-random $x$, this shows $(iii)$.
$h_x(W,D)\leq\eps'_x$ can also be shown.
\qed

\paragraph{Speed of convergence}
The main convergence Theorem~\ref{thMLconv} now immediately
follows from Propositions~\ref{proDtomu} and~\ref{proWtoD}. We
briefly remark on the convergence rate.
Lemma~\ref{lemHBounds} shows that $\E[\sum_t h_t(X,\mu)]$ is
logarithmic in the index $k_0$ of $\mu$ for $X=M$ ($\ln
w_{k_0}^{-1}\eqm\ln k_0$), but linear for $X=[W,D,\delta_{k_0}]$
($\ln\eps_{k_0}\eqm k_0)$.
The individual bounds for $\sum_t h_t(\hat\delta_{k_0},\mu)$ and
$\sum_t h_t(\hat\delta_{k_0},\hat D)$ in Proposition
\ref{proDtomu} are linear and exponential in $k_0$, respectively.
For $W\stackrel{M.L.}\longrightarrow D$ we could not establish any
convergence speed.

Finally we show that $W$ does not dominate all enumerable
semimeasures, as the definition of $W$ suggests. We summarize all
computability, measure, and dominance properties of $M$, $D$,
$\hat D$, and $W$ in the following theorem:

\ftheorem{thMWDprop}{Properties of $M$, $W$, $D$, and $\hat D$}{\hspace{1ex} \\
$(i)$ $M$ is an enumerable semimeasure, which dominates all enumerable semimeasures. %
$M$ is not computable and not a measure. \\
$(ii)$ $\hat D$ is a measure, $D$ is proportional to a measure, both
dominating all enumerable quasimeasures. %
$D$ and $\hat D$ are not computable 
and do not dominate all enumerable semimeasures. \\
$(iii)$ $W$ is an enumerable semimeasure, which dominates all enumerable quasimeasures. %
$W$ is not itself a quasimeasure, is not computable, and does not dominate all enumerable semimeasures.
}

We conjecture that $D$ and $\hat D$ are not even approximable
(limit-computable), but lie somewhere higher in the arithmetic
hierarchy. Since $W$ can be normalized to an approximable measure
M.L.-converging to $\mu$, and $D$ was only an intermediate
quantity, the question of approximability of $D$ seems not
too interesting.

\section{Conclusions}\label{secDisc}

We investigated a natural strengthening of Solomonoff's famous
convergence theorem, the latter stating that with probability 1
(w.p.1) the posterior of a universal semimeasure $M$ converges to
the true computable distribution $\mu$
($M\stackrel{w.p.1}\longrightarrow\mu$).
We answered partially negative the question of whether convergence
also holds individually for all Martin-L{\"o}f (M.L.) random
sequences
($\exists M : M${\scriptsize\tabcolsep=2pt
\begin{tabular}{c} \\[-12pt] $M.L.$ \\[-3pt] $\not\!\!\longrightarrow$
\end{tabular}}$\mu$).
We constructed random sequences $\a$ for which there exist
universal semimeasures on which convergence fails.
Multiplicative dominance of $M$ is the key property to show
convergence w.p.1. Dominance over all measures is also satisfied
by the restricted mixture $W$ over all quasimeasures. We showed
that $W$ converges to $\mu$ on all M.L.-random sequences by
exploiting the incomputable mixture $D$ over all measures. For
$D\stackrel{M.L.}\longrightarrow\mu$ we achieved a (weak)
convergence rate; for $W\stackrel{M.L.}\longrightarrow D$ and
$W/D\stackrel{M.L.}\longrightarrow 1$ only an asymptotic result.
The convergence rate properties w.p.1.\ of $D$ and $W$ are as
excellent as for $M$.

We do not know whether $D/\mu\stackrel{M.L.}\longrightarrow 1$
holds. We also don't know the convergence rate for
$W\stackrel{M.L.}\longrightarrow D$, and the current bound for
$D\stackrel{M.L.}\longrightarrow\mu$ is double exponentially worse
than for $M\stackrel{w.p.1}\longrightarrow\mu$. A minor question
is whether $D$ is approximable (which is unlikely).
Finally there could still exist {\em universal} semimeasures $M$
(dominating all enumerable semimeasures) for which
M.L.-convergence holds ($\exists M :
M\stackrel{M.L.}\longrightarrow\mu\,?$). In case they exist, we
expect them to have particularly interesting additional structure
and properties.
While most results in algorithmic information theory are
independent of the choice of the underlying universal Turing
machine (UTM) or universal semimeasure (USM), there are also
results which depend on this choice. For instance, one can show
that $\{(x,n):K_U(x)\leq n\}$ is tt-complete for some $U$, but
not tt-complete for others \cite{Muchnik:02}. A
potential $U$ dependence also occurs for predictions based on
monotone complexity \cite{Hutter:03unimdl}.
It could lead to interesting insights to identify a class of
``natural'' UTMs/USMs which have a variety of favorable
properties. A more moderate approach may be to consider classes
${\cal C}_i$ of UTMs/USMs satisfying certain properties ${\cal
P}_i$ and showing that the intersection $\cap_i {\cal C}_i$ is not
empty.

Another interesting and potentially fruitful approach to the
convergence problem at hand is to consider other classes of
semimeasures $\M$, define mixtures $M$ over $\M$, and (possibly)
generalized randomness concepts by using this $M$ in Definition
\ref{defML}. Using this approach, in \cite{Hutter:03unipriors} it
has been shown that convergence holds for a subclass of Bernoulli
distributions if the class is dense, but fails if the class is
gappy, showing that a denseness characterization of $\M$ could be
promising in general.

\paragraph{Acknowledgements}
We want to thank Alexey Chernov for his invaluable help.

\ifconf\else\newpage\fi
\addcontentsline{toc}{section}{References}
\begin{small}

\end{small}

\end{document}